\documentclass[final]{cvpr}

\usepackage{times}
\usepackage{epsfig}
\usepackage{graphicx}
\usepackage{amsmath}
\usepackage{amssymb}

\usepackage{multirow}
\usepackage{multicol}

\usepackage{balance}

\usepackage{array}
\newcolumntype{L}[1]{>{\raggedright\arraybackslash}m{#1}}
\newcolumntype{C}[1]{>{\centering\arraybackslash}p{#1}}

\usepackage{pifont}% http://ctan.org/pkg/pifont
\newcommand{\cmark}{\ding{51}}%
\newcommand{\xmark}{\ding{55}}%

\usepackage[pagebackref=true,breaklinks=true,colorlinks,bookmarks=false]{hyperref}

\def\cvprPaperID{WAD 4} % *** Enter the CVPR Paper ID here
\def\confYear{CVPR 2021}

\begin{document}

%%%%%%%%% TITLE
\title{Rethinking of Radar's Role: A Camera-Radar Dataset and \\Systematic Annotator via Coordinate Alignment}

\author{Yizhou Wang$^\text{1}$, Gaoang Wang$^\text{2}$, Hung-Min Hsu$^\text{1}$, Hui Liu$^\text{1,3}$, Jenq-Neng Hwang$^\text{1}$ \vspace{0.1em}\\
$^\text{1}$University of Washington, Seattle, WA, USA\\
$^\text{2}$Zhejiang University, Hangzhou, China\\
$^\text{3}$Silkwave Holdings Limited, Hong Kong, China\\
{\tt\small \{ywang26,hmhsu,huiliu,hwang\}@uw.edu, gaoangwang@intl.zju.edu.cn}
% For a paper whose authors are all at the same institution,
% omit the following lines up until the closing ``}''.
% Additional authors and addresses can be added with ``\and'',
% just like the second author.
% To save space, use either the email address or home page, not both
% \and
% Second Author\\
% Institution2\\
% First line of institution2 address\\
% {\tt\small secondauthor@i2.org}
}

\maketitle
% \thispagestyle{empty}% Reset page style to 'empty'

%%%%%%%%% ABSTRACT
\begin{abstract}
Radar has long been a common sensor on autonomous vehicles for obstacle ranging and speed estimation. However, as a robust sensor to all-weather conditions, radar's capability has not been well-exploited, compared with camera or LiDAR. Instead of just serving as a supplementary sensor, radar's rich information hidden in the radio frequencies can potentially provide useful clues to achieve more complicated tasks, like object classification and detection.  
In this paper, we propose a new dataset, named CRUW\footnote{Dataset available at \url{https://www.cruwdataset.org/}.}, with a systematic annotator and performance evaluation system to address the radar object detection (ROD) task, which aims to classify and localize the objects in 3D purely from radar's radio frequency (RF) images. To the best of our knowledge, CRUW is the first public large-scale dataset with a systematic annotation and evaluation system, which involves camera RGB images and radar RF images, collected in various driving scenarios.  
\end{abstract}

%%%%%%%%% BODY TEXT
\section{Introduction}
\label{sec:introduction}

\begin{figure}[t]
    \begin{center}
    \includegraphics[width=\linewidth]{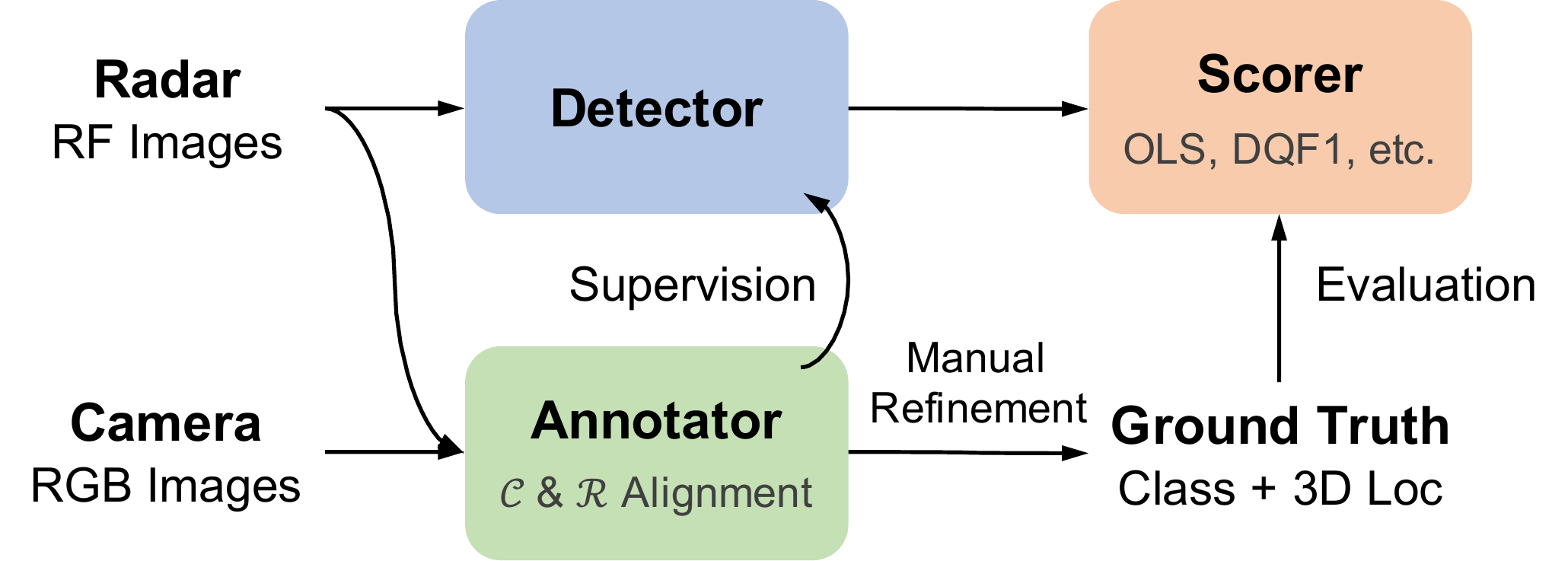}
    \end{center}
    \caption{The pipeline of the radar object detection (ROD) task, including a detector, an annotator, and a scorer. The detector takes radar data as the only input and can predict object classes and 3D locations in adverse driving scenarios. The annotator manages to align detections between camera $\mathcal{C}$ and radar $\mathcal{R}$ coordinates to serve as a ``teacher'' during training the ``student'' detector. The ground truth of the testing set for evaluation purpose is further refined manually from annotator's results. Finally, a scorer, with defined evaluation metrics, is used on the testing set to evaluate the performance. }
    \label{fig:intro}
\end{figure}

Multi-modality data analytics is greatly involved in the autonomous or assisted driving systems \cite{levinson2011towards,yang2020hybrid,yang2021fast} to improve the robustness of object perception \cite{ren2015faster,wang2019exploit,hsu2020traffic,yang2020novel} in a variety of different driving scenarios. 
Among the common sensors, i.e., camera, LiDAR, radar, on the autonomous vehicles, the RGB images and point cloud data from cameras and LiDAR are relatively easy for human to understand since the semantic information they convey is obvious. For example, 2D and 3D bounding boxes are intuitive for human to annotate the objects from RGB images and LiDAR point clouds, respectively. 
Therefore, some large and well-labeled datasets \cite{geiger2013vision,nuscenes2019,waymo_open_dataset,apollo_scape_dataset} have been released in the autonomous driving community to help develop and validate the machine learning algorithms. 
As an accurate 3D sensor for autonomous vehicles, LiDAR still faces the following critical limitations: 1) LiDAR is usually equipmental complex and computational expensive, so that not suitable for common industry use; 2) Laser transmitted by the LiDAR is not robust to occlusion or adverse weather scenarios. 
Radar, on the other hand, is a reliable and cost-efficient sensor capturing reliable 3D information even under adverse driving conditions, e.g., strong/weak lighting or bad weather. It is often used as supplement for other sensors due to its difficulty in parsing useful clues for semantic understanding. But this data unintuitiveness does not mean that radar has low potentials. 
% Therefore, the camera and radar become a great compensation sensor pair \textit{without LiDAR}, where camera can provide rich semantic information and high resolution, while radar can give the accurate and robust 3D ranging and velocity information. 

% possibility of radar to achieve object detection
The use of the frequency modulated continuous wave (FMCW) radar in most autonomous driving solutions lies in its obstacle ranging and speed estimation. The capability of radar's semantic understanding, e.g., object classification and detection, has not been well-exploited. The feasibility of achieving this semantic understanding owing to the hidden \textit{\textbf{phase}} information inside the radio frequencies. Typically, radar's amplitude is commonly used to estimate the distance and speed of the obstacles, while the phase information is usually not well-utilized because of its ``non-intuitiveness'', making it difficult to be handled by the classical signal processing mechanisms. 
There are two kinds of data representations for the FMCW radar, i.e., radio frequency (RF) image and radar points (see examples in the supplementary document). 
RF image is a much denser and more informative data representation, which requires further processing to understand the contents, containing both amplitude and phase information, but the location and speed are implicit. Whereas radar points are a kind of handy representation, which are usually sparse (less than 5 points on a nearby car) \cite{nuscenes2019, feng2020deep} and non-descriptive. 

% why need annotation? / why point-based?
Therefore, to extract those hidden features from RF images for semantic understanding, some researchers start to take advantage of the recent deep convolution neural networks (CNNs) to explore the possibility of radar-only object detection \cite{major2019vehicle,dong2020probabilistic,wang2021rodnet}, which usually require a large annotated dataset for training. However, radar data are very difficult to understand,  making human annotations significantly expensive or sometimes impossible to obtain. 
Besides, due to the low angular resolution of the common FMCW radar sensors, resulting in unreliable object dimension information from radar; more specifically, the bounding boxes defined in camera-based object detection are rarely used in the RF images, especially when the absence of LiDAR. Consequently, people usually represent objects as \textit{\textbf{points}} in the radar's bird's-eye view (BEV) coordinates instead \cite{wang2021rodnet}. These points are the reflection points of the radar signals from the obstacles in the radar's field of view (FoV). 
All of the above make the large-scale dataset, annotation and performance evaluation for object detection in the RF images very challenging while critically needed. 

% \begin{figure}[t]
%     \begin{center}
%     \includegraphics[width=\linewidth]{figures/data_eg.pdf}
%     \end{center}
%     \caption{Examples for RGB image, RF image, and radar points. (a) Three cars in the RGB image; (b) Annotation is extremely difficult on the corresponding RF image; (c) Radar points are sparse but have relative precise locations.}
%     \label{fig:data_eg}
% \end{figure}

In this paper, we propose a platform, including a large-scale dataset, a systematic annotation system, and a set of evaluation metrics for ROD, as shown in Fig.~\ref{fig:intro}.  
The proposed dataset contains about 400K synchronized camera-radar frames (3.5 hours) in various driving scenarios, i.e., parking lot, campus road, city street, and highway. As mentioned above, the radar data format is RF image with rich radio frequency information.
The proposed object annotation method calculates the optimal camera-radar bilateral coordinate projection and aligns the detections between the two coordinates. 
% The pipeline for our proposed annotation method is shown in Fig.~\ref{fig:pipeline}.
Two different kinds of detections are fed into the system, including object detection from camera and peak detection from radar. Then, a coordinate alignment strategy is utilized based on the proposed bilateral coordinate projection between camera and radar, and the ground plane for each frame is optimized by the alignment cost. 
In order to evaluate the performance of ROD, which predicts objects as points, we introduce a point-based similarity metric, called object location similarity (OLS), to serve as a matching score between two points in an RF image.
Based on OLS, to reasonably reflect the quality of the object detection results in the RF images, we introduce a series of evaluation metrics, considering object localization error, precision, and recall. Moreover, we define a new metric Detection Quality F1 Score (DQF1) that can jointly consider the above three metrics into a single metric to provide a comprehensive measurement of the detection performance. Unlike the widely used average precision (AP) and average recall (AR) defined for object detection tasks that emphasize classification, the proposed DQF1 focuses more on the localization accuracy.

Overall, the main contributions of this paper can be summarized as follows,
\begin{itemize}
    \item A large-scale dataset with synchronized camera-radar frames in various driving scenarios, including RF images as the radar data format for radar semantic understanding tasks. 
    \item An accurate and robust radar object annotator, that can systematically generate object labels for RF images, fusing the rich semantic information from a camera. It is also an object detector based on a camera-radar fusion manner in the normal driving scenarios.
    \item Derive the bilateral coordinate projection (BCP) between camera pixel coordinates and radar range-azimuth coordinates through the ground plane.
    \item Introduce a set of scoring metrics to evaluate the quality of ROD results comprehensively, including the proposed metrics, i.e., object location similarity (OLS) and detection quality F1 score (DQF1).
\end{itemize}

% The rest of this paper is organized as follows. Related works are presented in Section~\ref{sec:relatedworks}. The proposed dataset is introduced in Section~\ref{sec:dataset}, the annotation method is described in Section~\ref{sec:crdac}, and the evaluation system is shown in Section~\ref{sec:evaluation}. In Section~\ref{sec:experiments}, we evaluate on the baseline methods and validate the annotation quality of the proposed annotation system. Finally, we draw a conclusion in Section~\ref{sec:conclusion}.

\section{Related Works}
\label{sec:relatedworks}

% In this section, the related works of the radar-only object detection system are introduced, including detectors and annotation methods. 

\textbf{Detectors.} 
As mentioned in Section~\ref{sec:introduction}, radar-only object detection cannot reliably accomplish, especially the object-class identification, with sparse and non-descriptive radar points input. Therefore, most related research use the RF images as the input format. However, detecting objects from the radar RF data is very challenging because the inherent semantic information is not as obvious as that in the RGB images. Traditionally, peak detection algorithms are adopted to find the objects in the radar field of view (FoV), such as the widely used Constant False Alarm Rate (CFAR) detection algorithm \cite{richards2005fundamentals}. A classifier is then appended to classify the object class \cite{6042174,8468324}. However, these algorithms usually result in a large number of \textit{false positives} because it cannot distinguish the reflections of objects from that of background. Besides, it cannot reliably provide the object class. Moreover, one object may give multiple CFAR detections, which are confusing. Recently, some new techniques for radar data are proposed. Major~\textit{et~al.}~\cite{major2019vehicle} propose an automotive radar based vehicle detection method trained by LiDAR. However, they only consider vehicles as the target object class, and the scenarios are mostly highways without noisy obstacles.
Palffy~\textit{et~al.}~\cite{palffy2020cnn} propose a radar based, single-frame multi-class object detection method. However, they only consider the data from a single radar frame, which does not involve the object motion information. 
Wang \textit{et~al.}~\cite{wang2021rodnet2} propose the RODNet with temporal inception layers to capture temporal features of different lengths, an M-Net is used to merge and extract Doppler information from multiple chirps, and temporal deformable convolution is used to handle object relative motion. 

\textbf{Datasets.} 
Datasets are important to validate the algorithms, especially for the deep learning based methods. Since the release of the first complete autonomous driving dataset, i.e., KITTI \cite{geiger2013vision}, larger and more advanced datasets are now available \cite{apollo_scape_dataset,waymo_open_dataset,nuscenes2019}. However, due to the hardware compatibility and less developed radar perception techniques, most datasets do not incorporate radar signals as a part of their sensor systems. 
Among the available radar datasets (summarized in Table~\ref{tab:datasets}), some of them \cite{nuscenes2019,meyer2019automotive,RadarRobotCarDatasetICRA2020,sheeny2020radiate} consider radar data in the format of radar points that do not contain the useful Doppler and surface texture information of objects. 
Later, researchers start to focus on RF images as the radar data format. More specifically, some manage to collect a dataset with camera, radar and LiDAR, and annotate the objects as 3D bounding boxes based on the dense point cloud from LiDAR \cite{major2019vehicle,dong2020probabilistic}. Others consider the camera-radar solution without a LiDAR \cite{ouaknine2020carrada,palffy2020cnn}, whose annotation format is usually in pixel or point level. However, most of the datasets with RF images are not publicly available except CARRADA \cite{ouaknine2020carrada}. But CARRADA only contains one simple and easy scenario, i.e., parking lot, and is not suitable for practical usage.

% After extensive research on the available datasets, we cannot find a suitable one that includes large-scale radar data in RF image format with labeled ground truth.

\begin{table*}[t]
\small
    \begin{center}
    \begin{tabular}{l|cccccccc}
        \hline
        Dataset & Year & Scale & Scenarios$^*$ & Radar Format & Classes & Anno Source & Anno Format & Public \\
        \hline
        nuScenes \cite{nuscenes2019} & 2019 & 5.5 hours & ML & Radar Points & 23 & LiDAR & 3D Box & \cmark \\
        Qualcomm \cite{major2019vehicle} & 2019 & 3 hours & HW & RF Images & 1 & LiDAR & 3D Box & \xmark \\
        Astyx HiRes2019 \cite{meyer2019automotive} & 2019 & 546 frames & Urban & Radar Points & 7$^\dagger$ & LiDAR & 3D Box & \cmark \\
        RadarRobotCar \cite{RadarRobotCarDatasetICRA2020} & 2020 & 280 km & Urban & Radar Points & 0 & -- & -- & \cmark \\
        CARRADA \cite{ouaknine2020carrada} & 2020 & 21.2 min & PL & RF Images & 3 & Camera & Pixel & \cmark \\
        Xsense.ai \cite{dong2020probabilistic} & 2020 & 34.2 min & HW & RF Images & 1 & LiDAR & 3D Box & \xmark \\ 
        RTCnet \cite{palffy2020cnn} & 2020 & 1 hour & Urban & RF Images & 3 & Camera & Point & \xmark \\
        RADIATE \cite{sheeny2020radiate} & 2020 & 3 hours & ML & Radar Points & 7 & LiDAR$^\mathsection$ & 2D Box & \cmark \\
        \hline
        \textbf{CRUW (Ours)} & 2021 & 3.5 hours & ML & RF Images & 3 & Camera & Point & \cmark \\
        \hline
    \end{tabular}
    \end{center}
    \caption{Related datasets with radar data. $^*$ML: multiple scenarios; HW: highway; PL: parking lot. $^\dagger$Significantly imbalanced object distribution where car is the majority class. $^\mathsection$Details are not mentioned in the paper.}
    \label{tab:datasets}
\end{table*}

\textbf{Annotation methods.}
When LiDAR is not available to provide reliable object annotations, many people manage to fuse the semantic information from cameras. There are some camera-based techniques that are helpful for the radar object annotation task. Camera-based object detection \cite{ren2015faster,he2017mask,cai2018cascade,redmon2018yolov3,lin2017focal,duan2019centernet} aims to detect every object with its class and precise bounding box location from RGB images. 
Besides, some recent works try to infer 3D information from the 2D RGB images.
Some methods \cite{mousavian20173d,murthy2017reconstructing} localize vehicles by estimating their 3D structures using a CNN. Others \cite{song2015joint,ansari2018earth} try to develop a real-time monocular structure-from-motion (SfM) system, taking into account different kinds of cues. However, the above methods only work for the vehicles, which can satisfy the rigid-body structure assumption. 
To overcome this limitation, Wang \textit{et al.}~\cite{wang2019monocular} propose an accurate and robust object 3D localization system, based on the detected and tracked 2D bounding boxes of objects, which can work for most common moving objects in the road scenes, such as cars, pedestrians, and cyclists. 
The limitation is its inability to reliably estimate the 3D location of objects due to the inaccurate depth maps and non-negligible car sizes. 
After that, \cite{wang2021rodnet} proposes a probabilistic camera-radar fusion (CRF) algorithm to jointly consider the object 3D localization results from both camera and radar. But this kind of late fusion methods may introduce the errors from the early stages nor consider the correlation between different objects in the same frame. 
Recently, Ouaknine \textit{et al.}~\cite{ouaknine2020carrada} propose a semi-automatic annotation approach on radar data of some simple parking lot scenarios. However, its sensitivity to tracking and clustering makes it not robust in complex driving scenarios. 
Therefore, in this paper, we propose an annotation system to overcome the above issues.

\section{CRUW Dataset}
\label{sec:dataset}

\begin{figure}[t]
\begin{center}
\includegraphics[width=.85\linewidth]{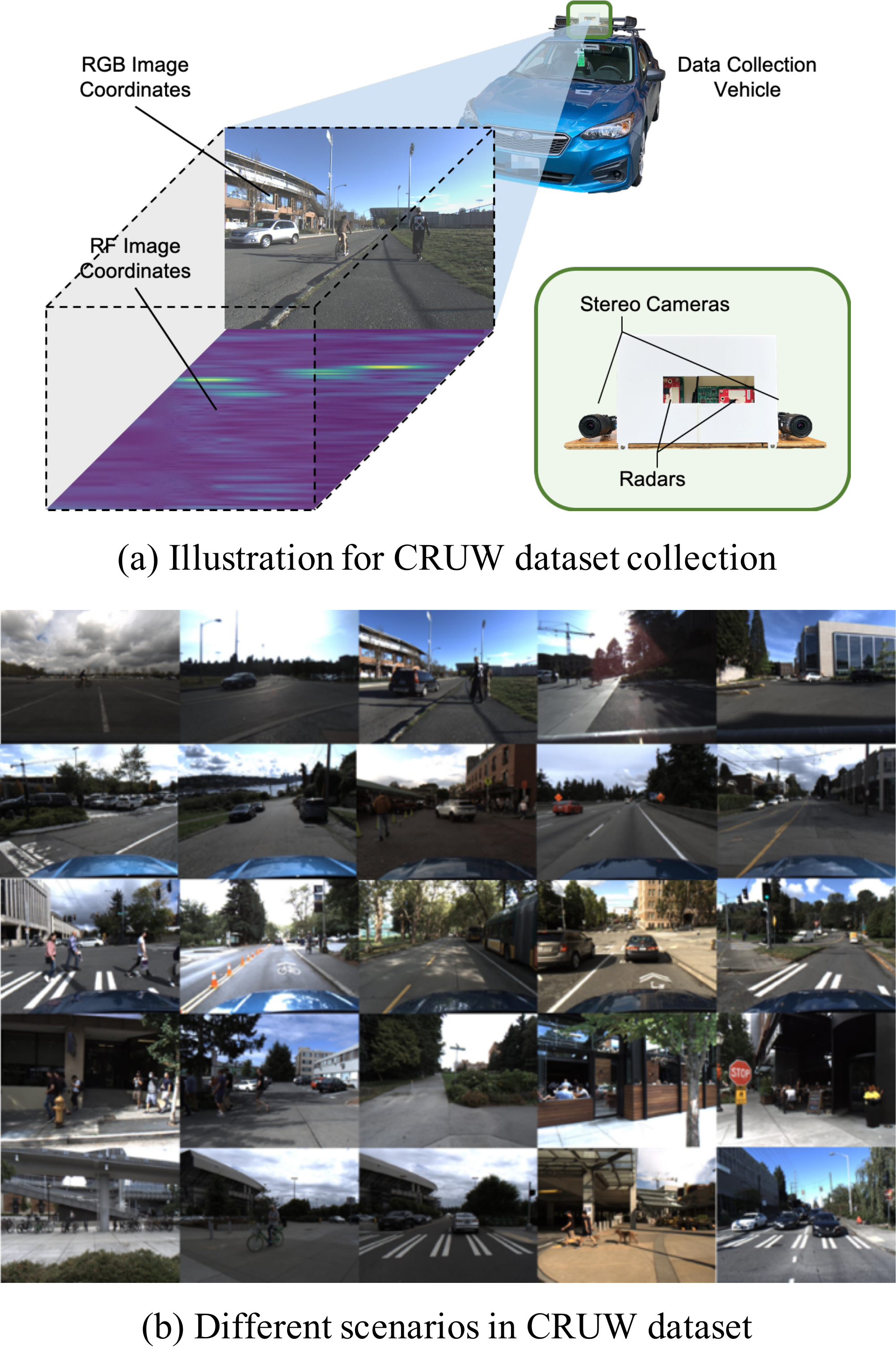}
\end{center}
  \caption{The sensor platform and some sample driving scenarios.}
  \label{fig:dataset}
\end{figure}

\subsection{Sensor System and Data Description}

\begin{table}[t]
    \footnotesize
    \begin{center}
    \begin{tabular}{l c | l c}
        \hline
        Camera & Value & Radar & Value \\
        \hline
        Frame rate & 30 FPS & Frame rate & 30 FPS \\
        Pixels (W$\times$H) & 1440$\times$1080 & Frequency & 77 GHz \\
        Resolution & 1.6 MP & \# of transmitters & 2 \\
        Field of View & 93.6$^{\circ}$ & \# of receivers & 4 \\
        Stereo Baseline & 0.35 m & \# of chirps per frame & 255 \\
        && Range resolution & 0.23 m \\
        && Azimuth resolution & $\sim$15$^{\circ}$ \\
        \hline
    \end{tabular}
    \end{center}
    \caption{Sensor Configurations for CRUW Dataset.}
    \label{tab:sensor_config}
\end{table}

The sensor platform contains a pair of stereo cameras \cite{flir} and two perpendicular 77GHz FMCW mmWave radar antenna arrays \cite{ti}. The sensors are assembled and mounted together as shown in Fig.~\ref{fig:dataset}~(a).  Some configurations of our sensor platform are shown in Table~\ref{tab:sensor_config}. 

The proposed dataset contains 3.5 hours with 30 FPS (about 400K frames) of camera-radar data in different driving scenarios, including parking lot, campus road, city street, and highway. Some sample scenarios are shown in Fig.~\ref{fig:dataset}~(b). The data are collected in two different views, i.e., driver front view and driver side view, to validate different perspective views for autonomous or assisted driving. Besides, we also collect several vision-hard sequences of poor image quality, i.e., weak/strong lighting, blur, etc. These data are only used in the testing set for evaluation purpose. 

\subsection{Data Distribution}

\begin{table}[t]
    \footnotesize
    \begin{center}
    \begin{tabular}{c c c c}
        \hline
        Scenarios & \# of Seqs & \# of Frames & Vision-Hard \% \\
        \hline
        Parking Lot & 124 & 106K & 15\% \\
        Campus Road & 112 & 94K & 11\% \\
        City Street & 216 & 175K & 6\% \\
        Highway & 12 & 20K & 0\% \\
        \hline
        Overall & 464 & 396K & 9\% \\
        \hline
    \end{tabular}
    \end{center}
    \caption{Driving scenarios statistics for CRUW dataset.}
    \label{tab:scenarios_distri}
\end{table}

\begin{figure}[t]
  \begin{center}
  \includegraphics[width=\linewidth]{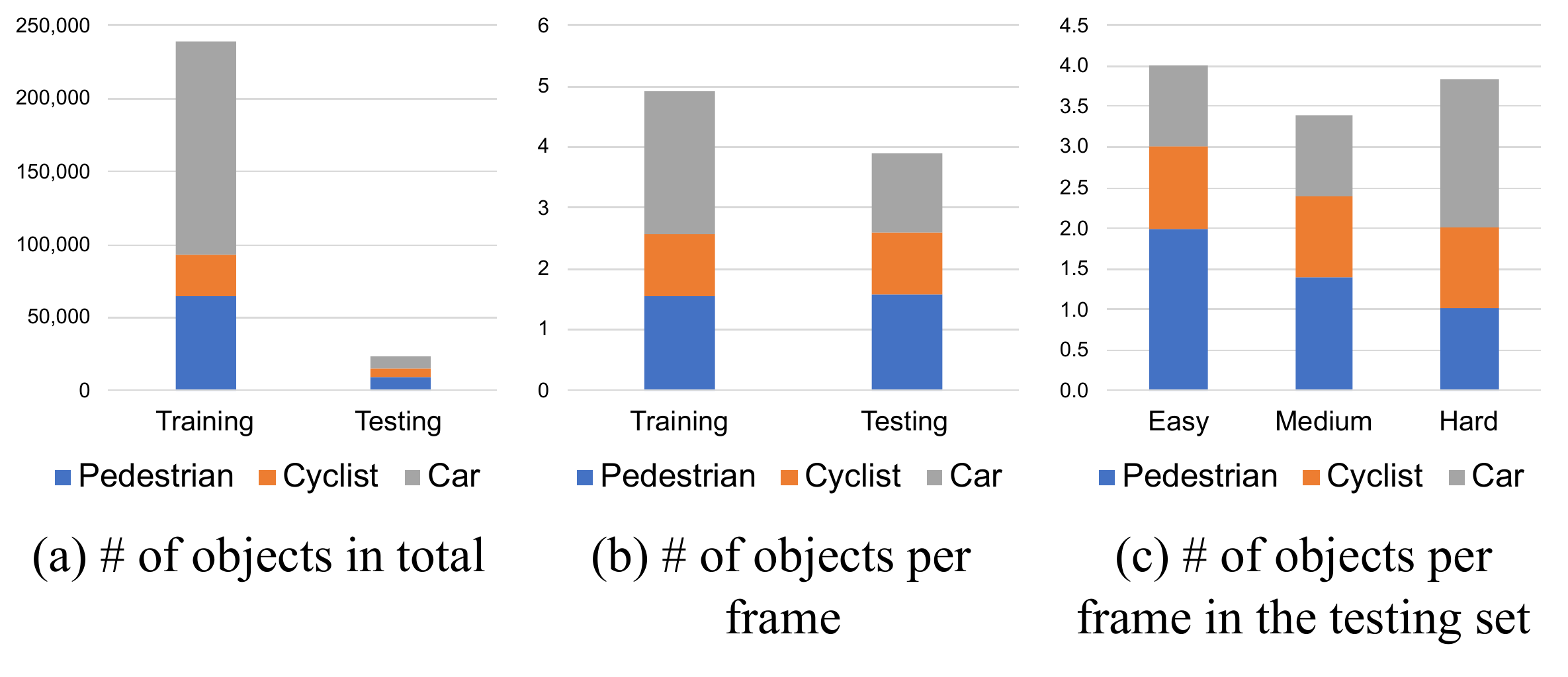}
  \end{center}
   \caption{Illustration for our CRUW dataset distribution. Here, (a)-(c) show the object distribution in the radar's FoV ($0$-$25$m, $\pm90^{\circ}$).}
\label{fig:cr_distribution}
\end{figure}

The data distribution is shown in Table~\ref{tab:scenarios_distri} and Fig.~\ref{fig:cr_distribution}. The object statistics in Fig.~\ref{fig:cr_distribution} only consider the objects within the radar's field of view (FoV), i.e., $0$-$25$m, $\pm 90^{\circ}$, based on the current hardware capability. There are about 260K objects in CRUW dataset in total, including $92\%$ for training and $8\%$ for testing. The average number of objects in each frame is similar between training and testing data. 

The four different driving scenarios, i.e., parking lot, campus road, city street, and highway, are shown in Table~\ref{tab:scenarios_distri} with the number of sequences, frames and vision-hard percentages. 
From each scenario, we randomly select several complete sequences as testing sequences, which are not used for training. Thus, the training and testing sequences are captured at different locations and different time.
For the ground truth needed for evaluation purposes, $10\%$ of the visible and $100\%$ of the vision-hard data are human-labeled by manually refinement on the results of the annotator introduced in Section~\ref{sec:crdac}.

\section{Radar Object Annotation System}
\label{sec:crdac}

\begin{figure}[t]
    \begin{center}
    \includegraphics[width=.8\linewidth]{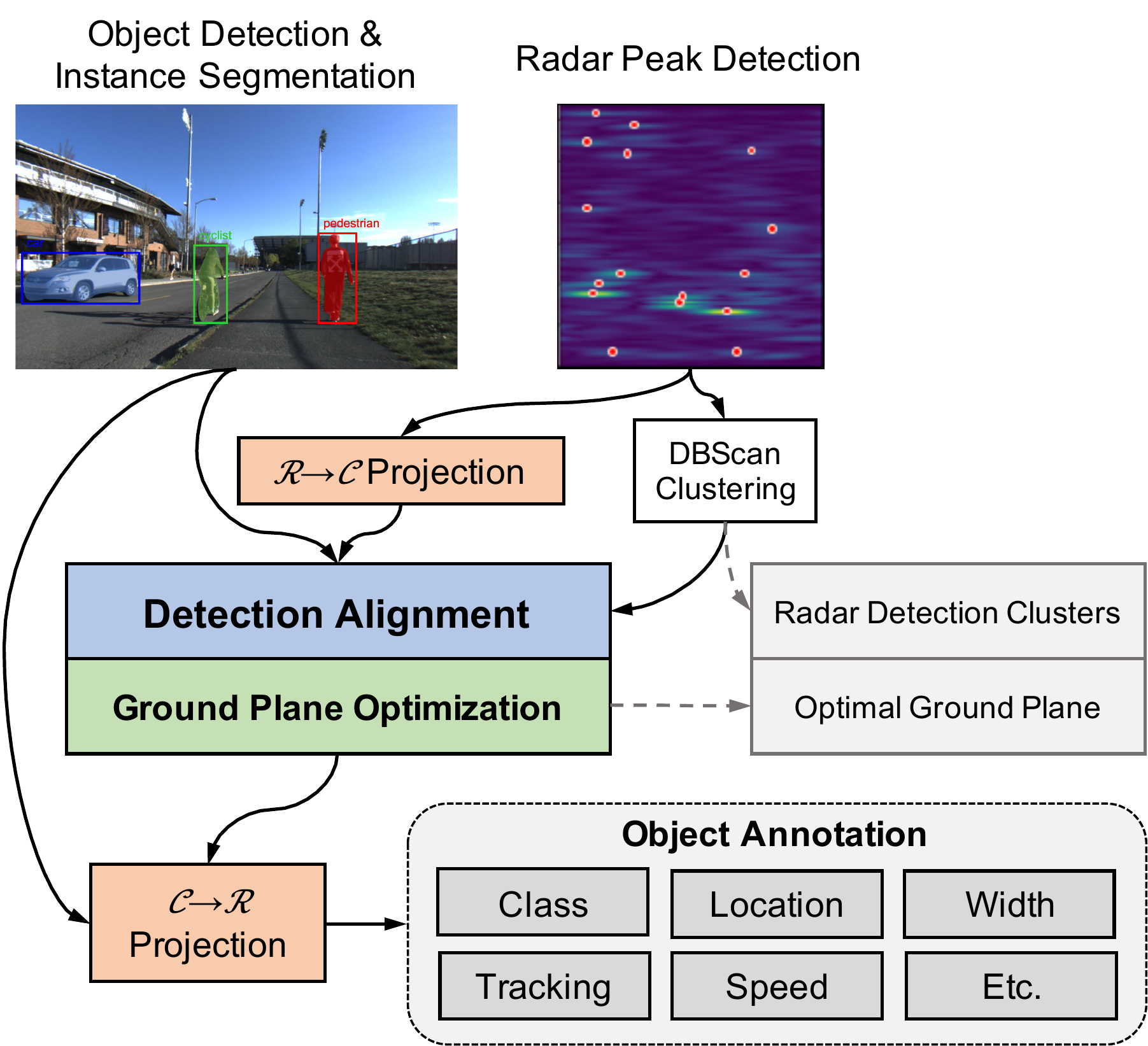}
    \end{center}
    \caption{The framework of our proposed annotation system for automatic radar object annotation. The system takes detections inferred from both RGB images and RF images, and manages to align the detections between the RGB image coordinates $\mathcal{C}$ and radar range-azimuth coordinates $\mathcal{R}$ through the corresponding ground plane. After detections are aligned, the system can provide reliable object annotation taking advantages of camera information, including object attributes, e.g., class, location, reflection width, tracking, speed, etc., as well as radar detection clusters and the optimal ground planes.}
    \label{fig:pipeline}
\end{figure}

In this section, our proposed radar object annotation system (Fig.~\ref{fig:pipeline}) will be introduced. First, the input data, both from camera and radar, are pre-processed and fed into our systematic annotation system for initialization. Second, the bilateral coordinate projection is derived to connect the camera and radar coordinates. Then, the detection alignment strategy with the ground plane optimization is applied to accurately detect and localize the objects in RF images. Overall, the radar detections are aligned and clustered by different instances, and the final object annotations are the centers of the resulting clusters.

\subsection{Notations}
\label{subsec:crdac_ps}

\textbf{Coordinates.} 
We first define two 2D coordinates used in our system: 1) Camera 2D pixel coordinates $(u, v) \in \mathcal{C}$; 2) Radar range-azimuth coordinates, i.e., bird's-eye view (BEV), $(r, \theta) \in \mathcal{R}$; 3) 3D camera coordinates with the origin at the camera center $(x^c, y^c, z^c) \in \mathcal{W}^{c}$; and 4) 3D radar coordinates with the origin at the radar $(x^r, y^r, z^r) \in \mathcal{W}^{r}$. 
The system takes two different kinds of detections from both camera and radar, i.e., object detection from RGB images and peak detection from RF images, as the input, and these detections are defined within $\mathcal{C}$ and $\mathcal{R}$. 

\begin{figure}[t]
    \begin{center}
    \includegraphics[width=.7\linewidth]{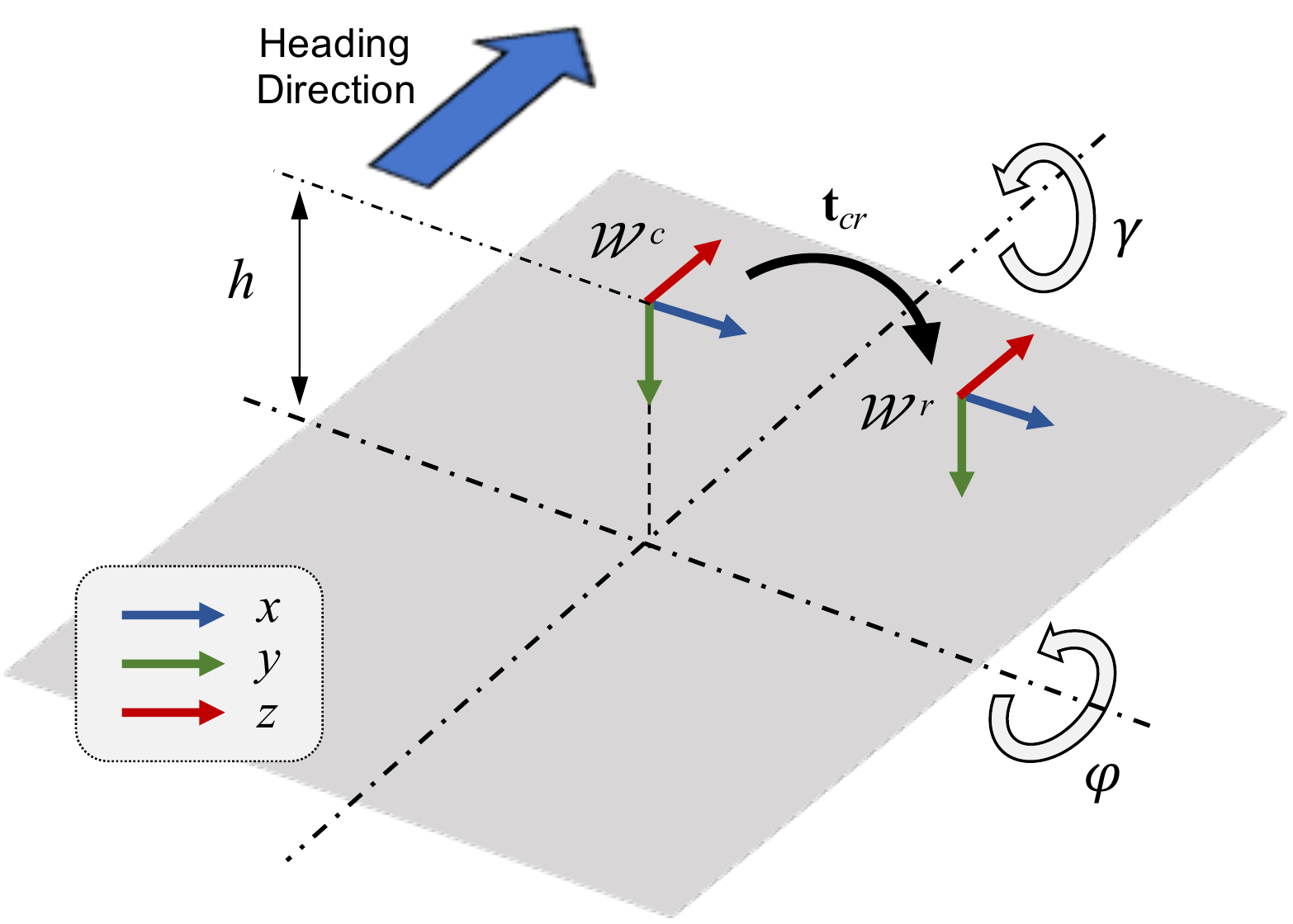}
    \end{center}
    \caption{Illustration for 3D coordinates and ground plane parameters. The gray rectangle represents the ground plane. The camera and radar mounted on a vehicle are above the ground plane. $\mathbf{t}_{cr}$ is the translation vector from $\mathcal{W}^c$ to $\mathcal{W}^r$.}
    \label{fig:coors}
\end{figure}

\textbf{Ground plane parameters.} 
We define the ground plane by three ground plane parameters, i.e., two rotation angles and one offset. The ground plane parameters for each frame can be represented as 
\begin{equation}
    \mathbf{g} = [\varphi, \gamma, h],
\end{equation}
where $\varphi$ and $\gamma$ respectively denote the pitch and roll angles of the ground plane w.r.t. the $z$-axis of $\mathcal{W}^{c}$; $h$ is the camera height. The illustration of these ground plane parameters is shown in Fig.~\ref{fig:coors}.

\subsection{Initialization}
\label{subsec:crdac_init}

In this work, we use RF images to represent our radar signal reflections. They are in radar 2D range-azimuth coordinates $(r, \theta) \in \mathcal{R}$, where the $\theta$-axis denotes azimuth (angle) and the $r$-axis denotes range (distance).
% For an FMCW radar, it transmits continuous chirps and receives the reflected echoes from the obstacles. After the echoes are received and processed, the fast Fourier transform (FFT) is performed on the samples to estimate the range of the reflections. A low-pass filter (LPF) is then utilized to remove the high-frequency noise across all chirps in each frame at the rate of 30 FPS. After the LPF, a second FFT is performed on the samples along different receiver antennas to estimate the azimuth angle of the reflections and obtain the final RF images. 
Two different kinds of detections from camera and radar are used as the input to our system: 
\begin{itemize}
    \item Object detection and instance segmentation from RGB images using a Mask R-CNN object detector \cite{he2017mask}. 
    % The network is pre-trained on the KITTI dataset \cite{geiger2013vision} and fine-tuned on the Cityscapes dataset \cite{cordts2016cityscapes}.
    \item Radar peak detection on RF images using the CFAR detection algorithm \cite{richards2005fundamentals}.
\end{itemize}

After the data and detections are prepared, we need to initialize the systematic annotation system. First, the ground plane parameters are initialized by the sensor calibration results. Specifically, for the CRUW dataset, $\varphi_0 = 4^{\circ}$, $\gamma_0 = 0^{\circ}$, $h_0 = 1.65$m. 
Second, the density-based spatial clustering of applications with noise (DBScan) clustering algorithm \cite{schubert2017dbscan} is implemented on the CFAR detections to obtain the initial detection clusters. 
% The DBScan clustering algorithm is frequently used to cluster the radar detection and decide which reflections are from the same object and which ones are background noises. The strength of adopting DBScan clustering is that it is a density-based spatial clustering algorithm and is robust to noises, which has very similar application scenario with radar detections. 
% The initial clustering result from DBScan is used in the following two detection alignment stages to make the alignment robust to the non-overlapped field of views (FoVs) between camera and radar. 

\subsection{Bilateral Coordinate Projection}
\label{subsec:crdac_bcp}

Consider the ground plane parameters defined in Section~\ref{subsec:crdac_ps}, we derive the camera-radar bilateral coordinate projection (BCP) for any point on the ground plane. We start from projecting a point $(r, \theta) \in \mathcal{R}$ to $\mathbf{x}^c = (x^c, y^c, z^c)$. 

We first do a polar to Cartesian coordinate transformation and transform to $\mathcal{W}^c$ by sensor calibration of the translation from camera to radar $\mathbf{t}_{cr} = [t_{cr,x}, t_{cr,y}, t_{cr,z}]^{\top}$,
\begin{equation}
\begin{aligned}
    x^c = r \sin (\theta) + t_{cr,x}, \\
    z^c = r \cos (\theta) + t_{cr,z}.
\end{aligned}
\label{eq:trans}
\end{equation}
Here, we ignore the rotation between $\mathcal{W}^r$ and $\mathcal{W}^c$ since both sensors are well-calibrated with the same orientation. 

\begin{figure}[t]
    \begin{center}
    \includegraphics[width=.95\linewidth]{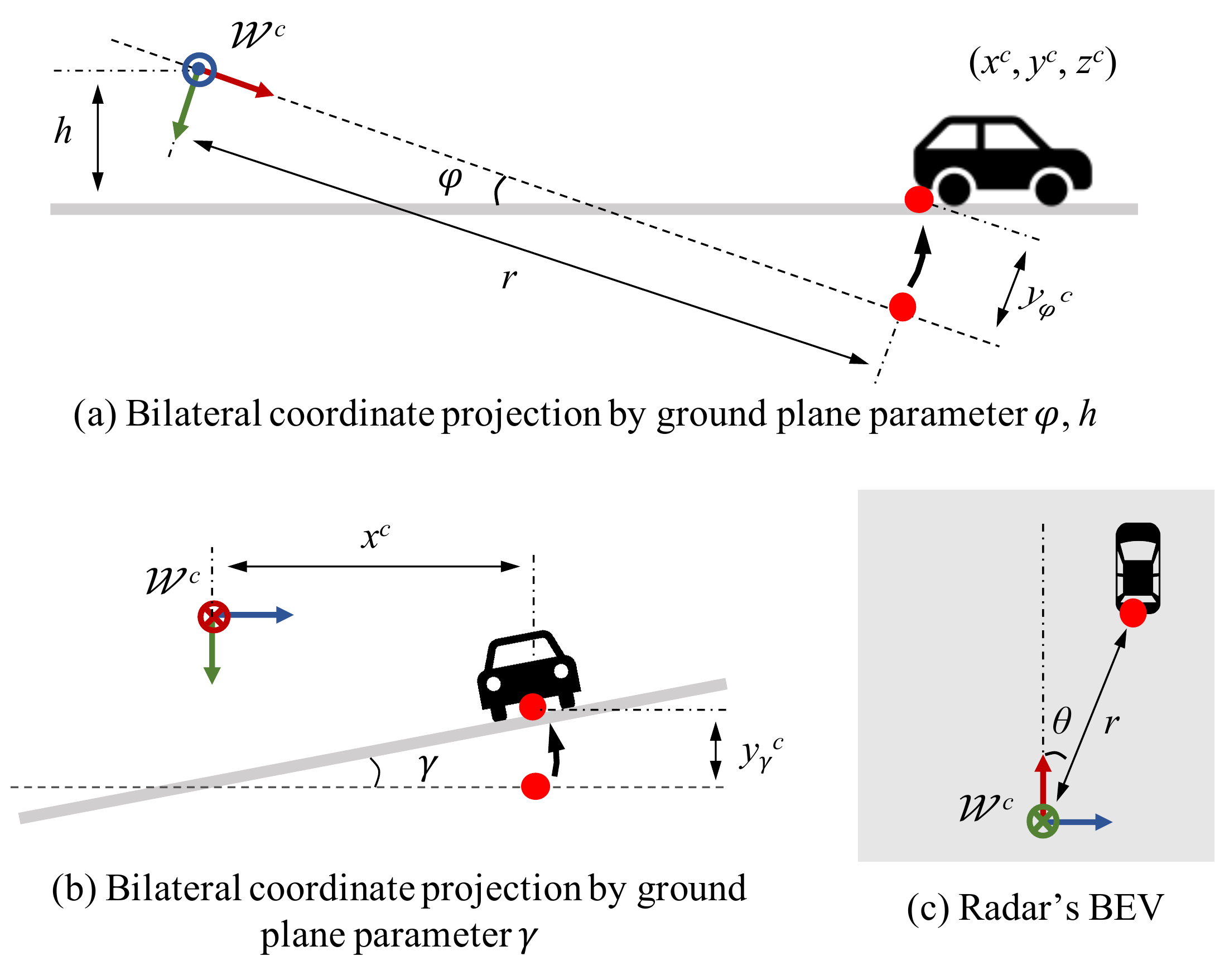}
    \end{center}
    \caption{Geometric illustrations for our proposed bilateral coordinate projection. The projection is split into two steps, i.e., $\varphi$ rotation and $\gamma$ rotation, where the gray lines represent the ground planes. To clearly show the projection, these two steps are shown in two different perspectives, where (a) is based on the parameters $\varphi$ and $h$; (b) is based on the parameter $\gamma$. (c) represents a car in the radar's BEV with range $r$ and azimuth angle $\theta$.}
    \label{fig:bcp}
\end{figure}

To calculate $y^c$, we first consider the ground plane with only pitch rotation angle $\varphi$. Assuming a small $\varphi$, which is a valid assumption in driving scenarios, $y^c_{\varphi}$ can be approximated by the similar triangles rule as 
\begin{equation}
\begin{aligned}
    y^c_{\varphi} \approx h - r \sin (\varphi).
\end{aligned}
\end{equation}
By taking the second rotation angle $\gamma$ into consideration, 
\begin{equation}
    y^c_{\gamma} = x^c \tan (\gamma).
\end{equation}
The illustrations for these two projections are shown in Fig.~\ref{fig:bcp} (a) and (b).
Therefore, the overall $y^c$ can be represented as
\begin{equation}
\begin{aligned}
    y^c &= y^c_{\varphi} - y^c_{\gamma} = h - r \sin (\varphi) - x^c \tan (\gamma).
\end{aligned}
\end{equation}

Finally, we project $\mathbf{x}^c \in \mathcal{W}^c$ to $\mathcal{C}$ by the camera intrinsic matrix $\mathbf{K}$.
Put it together, we can project a point $(r, \theta) \in \mathcal{R}$ to $(u, v) \in \mathcal{C}$ by the following $\mathcal{R} \rightarrow \mathcal{C}$ projection, denoted as $P(\cdot)$,
\begin{equation}
    [u, v]^{\top}
    = P(r, \theta),
    \label{eq:r2c}
\end{equation}
where
\begin{equation}
    \begin{aligned}
    u &= f_x \frac{r \sin (\theta) + t_{cr,x}}{r \cos (\theta) + t_{cr,z}} + c_x, \\
    v &= f_y \frac{h - r \sin (\varphi) - (r \sin (\theta) + t_{cr,x}) \tan (\gamma)}{r \cos (\theta) + t_{cr,z}} + c_y.
    \end{aligned}
    \label{eq:r2c_detail}
\end{equation}
Here, $(f_x, f_y)$ represents the camera's focal length; $(c_x, c_y)$ represents the camera center.

After the $\mathcal{R} \rightarrow \mathcal{C}$ projection is properly derived, we would like to consider the other direction, i.e., $\mathcal{C} \rightarrow \mathcal{R}$ projection. Since the projection in Eq.~\ref{eq:r2c} is a 2D-to-2D projection without losing any degrees of freedom (DoF), it can be reversed. Therefore, we define the $\mathcal{C} \rightarrow \mathcal{R}$ projection as
\begin{equation}
    [r, \theta]^{\top}
    = P^{-1} (u, v).
    \label{eq:c2r}
\end{equation}
Here, the $\mathcal{C} \rightarrow \mathcal{W}^c$ projection can be derived as
\begin{equation}
    \begin{aligned}
    x^c &= \frac{h\hat{x}}{\sqrt{1+\hat{x}^2} \sin(\varphi) + \hat{x} \tan(\gamma) + \hat{y}}, \\
    z^c &= \frac{h}{\sqrt{1+\hat{x}^2} \sin(\varphi) + \hat{x} \tan(\gamma) + \hat{y}},
    \end{aligned}
\end{equation}
where
\begin{equation}
    \begin{aligned}
    \hat{x} = \frac{x^c}{z^c} = \frac{u - c_x}{f_x}, \\
    \hat{y} = \frac{y^c}{z^c} = \frac{v - c_y}{f_y}.
    \end{aligned}
\end{equation}
% Note that $\mathcal{W}^c \rightarrow \mathcal{R}$ projection just contains the camera to radar translation and a Cartesian to polar coordinate transformation, which is trivial and will not be elaborated in detail.

% Therefore, we define the BCP between camera and radar using Eq.~\ref{eq:r2c} and Eq.~\ref{eq:c2r}, so that we can align these two coordinates by the ground plane parameters. 

\subsection{Detection Alignment and Optimization}
\label{subsec:crdac_cfda}

Based on the BCP, the connection between camera and radar is established through the ground plane parameters. To align these two coordinates, we come up with a detection alignment strategy by ground plane optimization. 

\textbf{Detection alignment cost.} 
The alignment between $\mathcal{R}$ and $\mathcal{C}$ is challenging because they are both 2D coordinates with significantly different perspectives. However, we find that the object height is very useful information during the alignment. The object will have very different heights at different distances in the RGB image due to the perspective. Thus, the object height is a special cue of its 3D location. 
In the alignment, we use the average height for each object class and project all the CFAR peak detections to RGB images by $\mathcal{R} \rightarrow \mathcal{C}$ projection with the object height. Each CFAR detection will be projected as a line segment in the RGB image, named as CFAR line. We define the detection alignment cost for a CFAR detection $i$ to be
\begin{equation}
    \ell_i = \lambda_i \left( h_{i} - h^{mask}_{i} \right)^2 + (1 - \lambda_i) \left( h_{i} - h^{bbox}_{i}\right)^2,
    \label{eq:align}
\end{equation}
where $h_i$ is the height of the projected CFAR detection in the RGB image; $h^{mask}_{i}$ and $h^{bbox}_{i}$ are the heights of the object mask and bounding box near the projected CFAR line, respectively. $\ell_i$ is the combined cost from object mask and bounding box based on an adaptive weight $\lambda_i$, where
\begin{equation}
    \lambda_i = e^{-\alpha z^c_{i}},
\end{equation}
which means $\lambda_i$ is dependent on $z^c_i$. The intuition behind is that, for nearby objects, 2D object masks can accurately describe the height of the objects, especially for the cars, whereas bounding boxes fail to do that. While for faraway objects, bounding boxes can perform better than the unreliable masks. Here, $\alpha$ is a parameter to distinguish between nearby and faraway objects. During the experiment, we empirically find $10$ meters is a good threshold, so that we set $\alpha = 0.06$ to make $\lambda = e^{-10\alpha} \approx 0.5$.

\textbf{Ground plane optimization.}
After the detections are aligned between camera and radar, the ground plane parameters can be optimized accordingly. 
Here, we define the objective function as
\begin{equation}
    \min_{\varphi, \gamma} \sum_{t} \sum_{i=1}^{n_{CFAR}} \left( v_{i, t} - v^{mask}_{i, t} \right)^2,
    \label{eq:opt}
\end{equation}
where $n_{CFAR}$ is the number of CFAR detections in the frame $t$; $v_{i, t}$ is the vertical pixel location of the CFAR detection, i.e., $v$-axis of the lower endpoint of the CFAR line; $v^{mask}_{i, t}$ is the bottom location of the object mask near the projected CFAR line.

When the ground plane is optimized for each frame, we can use the $\mathcal{C} \rightarrow \mathcal{R}$ projection to project the detections in the RGB image that are not aligned. Here, we call this procedure \textit{supplementary projection}. Since the camera usually can detect more objects than radar, this projection can significantly improve the detection recall.

\begin{table*}[t]
\begin{center}
\small
\begin{tabular}{l|c|ccc|ccc}
\hline
Method & Scenario & MAE & Precision & Recall & \textbf{AP} & \textbf{AR} & \textbf{DQF1}\\
\hline
\multirow{4}{*}{RODNet (Vanilla) \cite{wang2021rodnet}} 
& Overall & 0.31 ($\pm$0.26) & 95.90\% & 78.03\% & 74.29\% & 77.85\% & 81.02\% \\
& Parking Lot & 0.26 ($\pm$0.19) & 98.29\% & 87.76\% & 85.33\% & 86.76\% & 89.33\% \\
& Campus Road & 0.42 ($\pm$0.30) & 89.49\% & 53.02\% & 42.67\% & 49.03\% & 56.03\% \\
& City Street & 0.48 ($\pm$0.39) & 88.88\% & 73.42\% & 59.79\% & 67.23\% & 71.15\% \\
\hline
\multirow{4}{*}{RODNet (HG) \cite{wang2021rodnet}} 
& Overall & 0.31 ($\pm$0.23) & 96.02\% & 88.56\% & 83.76\% & 85.62\% & 86.64\% \\
& Parking Lot & 0.26 ($\pm$0.16) & 98.26\% & 96.94\% & 93.60\% & 94.98\% & 93.63\% \\
& Campus Road & 0.40 ($\pm$0.26) & 92.16\% & 68.76\% &50.34\% & 57.23\% & 70.28\% \\
& City Street & 0.48 ($\pm$0.39) & 91.53\% & 81.27\% & 64.54\% & 70.47\% & 75.55\% \\
\hline
\multirow{4}{*}{RODNet (Full) \cite{wang2021rodnet2}} 
& Overall & 0.31 ($\pm$0.25) & 95.93\% & 88.86\% & 85.98\% & 87.86\% & 87.82\% \\
& Parking Lot & 0.27 ($\pm$0.21) & 98.49\% & 97.98\% & 95.79\% & 96.85\% & 94.62\% \\
& Campus Road & 0.36 ($\pm$0.26) & 92.08\% & 69.40\% & 57.06\% & 62.08\% & 73.62\% \\
& City Street & 0.49 ($\pm$0.37) & 91.59\% & 76.37\% & 62.83\% & 70.41\% & 74.65\% \\
\hline
\end{tabular}
\end{center}
\caption{Performance evaluation using our proposed scoring metrics for a radar-only object detection method (RODNet) on the \textit{\textbf{testing}} set under different driving scenarios. }
\label{tab:results_rodnet}
\end{table*}

\section{Point-based Detection Evaluation}
\label{sec:evaluation}

As mentioned in Section~\ref{sec:introduction}, the common datasets with the FMCW radar  \cite{nuscenes2019,RadarRobotCarDatasetICRA2020,ouaknine2020carrada,sheeny2020radiate,wang2021rodnet} usually use points to represent detections, we propose an evaluation system for point-based detection without any bounding box. 

\subsection{Point-based Similarity}

Before introducing the evaluation metrics, the matching strategy between the generated annotations and the ground truth needs to be explained. 
We propose the object location similarity (OLS) to represent the similarity between a detection $i$ and a ground truth $j$ to be
\begin{equation}
    \text{OLS}(i,j) = \exp \left\{ \frac{-d_{ij}^2}{2 (s_j \kappa_{cls})^2} \right\},
    \label{eq:ols}
\end{equation}
where $d$ is the distance (in meters) between the two points in an RF image; $s$ is the object distance from the sensors; and $\kappa_{cls}$ is a per-class constant that represents the error tolerance for class $cls$, which can be determined by the object average size of the corresponding class. 
% Since OLS is reasonably distributed between 0 and 1, we treat it as a good representation of the localization error, and use it as the matching threshold for the following evaluation metrics, i.e., $i$ and $j$ are matched if OLS$(i,j) > 0.5$.

% \begin{figure}
%     \begin{center}
%     \includegraphics[width=.7\linewidth]{figures/ols.pdf}
%     \end{center}
%     \caption{Illustration of the proposed point-based similarity OLS between a detection $i$ and a ground truth $j$.}
%     \label{fig:ols}
% \end{figure}

\begin{figure*}[t]
    \begin{center}
    \includegraphics[width=\linewidth]{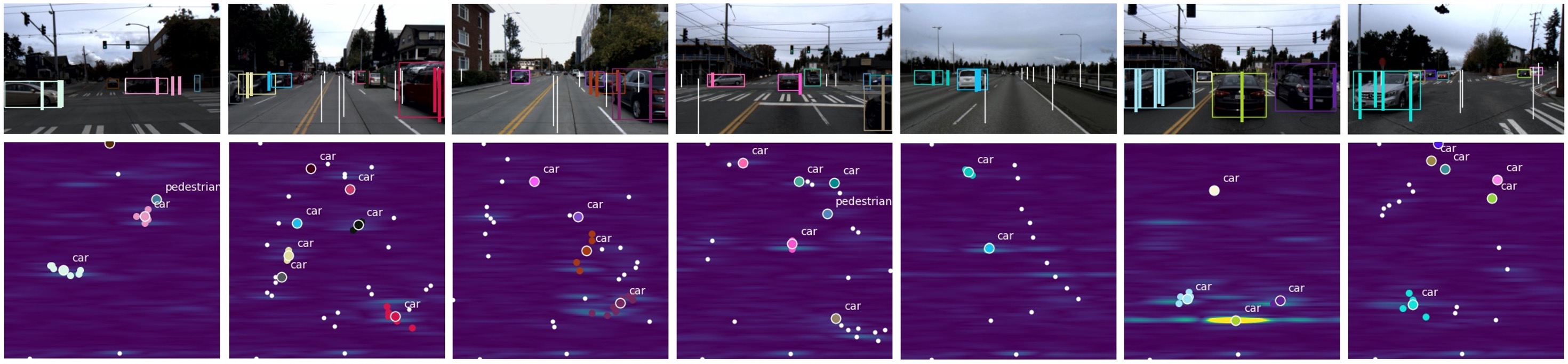}
    \end{center}
    \caption{Qualitative results for our proposed annotation system in various driving scenarios. The upper row shows the RGB images with the detected bounding boxes from Mask R-CNN and the projected CFAR detections (vertical lines). The lower row shows the RF images with the CFAR detections (dots) and the final object annotations. The colors of the detections illustrate the detection alignment results, whereas the outliers, i.e., background detections, are presented using the white lines and dots in RGB and RF images, respectively.}
    \label{fig:results}
\end{figure*}

\subsection{Scoring Metrics}
\label{subsec:metrics}

The quality of the radar object detection includes two aspects: 1) The location accuracy of the object detections; 2) The object-class correctness in the radar's FoV.

Typically, to evaluate the location accuracy, the \textit{mean absolute error} (MAE) is used to calculate the absolute localization error in meters. But MAE cannot reflect the false positives or false negatives, i.e., wrong or missing detections. Therefore, we also include \textit{precision} and \textit{recall} into our evaluation metrics. 

Moreover, to better describe the quality of the detections, we define a new evaluation metric called \textit{Detection Quality F1 Score} (DQF1), that aims to jointly consider MAE, precision and recall into one single number of measurement. Considering the F1 score that is frequently used to combine the precision and recall, we define DQF1 as
\begin{equation}
    \text{DQF1} = \frac{2}{n_{det} + n_{gt}} \sum_{j=1}^{n_{gt}} \sum_{i=1}^{n_{det}}  \delta_{i,j} \cdot \text{OLS}(i, j),
\end{equation}
where $n_{det}$ is the number of detections; $n_{gt}$ is the number of ground truth; $\delta_{i,j}$ is a binary flag to illustrate whether the $i$-th detection is matched with the $j$-th ground truth. 
Note that we use the summation of OLS to replace the number of true positives, so that the object localization accuracy is also involved. Since OLS is between 0 and 1, DQF1 is also a metric between 0 and 1.

\section{Baseline Evaluation}
\label{sec:experiments}

\subsection{Evaluation on Radar Object Detection}

We first evaluate the radar-only object detection performance of the state-of-the-art method, called RODNet \cite{wang2021rodnet,wang2021rodnet2}, using our proposed evaluation system. Vanilla, Hourglass (HG), and Full are three different network configurations mentioned in the original paper. The evaluation scores are shown in Table~\ref{tab:results_rodnet}. 
The comprehensive metrics for radar object detection include AP, AR, and the proposed DQF1. AP and AR used in the original paper are classical metrics for object detection, while DQF1 focuses more on localization accuracy instead of classification. 

It is obvious that the localization accuracy of the RODNet is very promising since it considers radar data as the only input. Therefore, the DQF1 scores are also a little bit higher than the original AP and AR, which shows the property of DQF1 that emphasize more on localization. 
Overall, with AP, AR, and DQF1, we can have an overall impression of the point-based detection performance from both classification and localization.

\subsection{Radar Object Annotation Comparison}

We use our proposed annotation system to automatically generate object annotations on CRUW dataset and evaluate the annotation quality by the evaluation metrics mentioned in Section~\ref{subsec:metrics}, on the selected crowded training set. Note that the MAE is evaluated by both mean and standard deviation of the localization errors.

We compare the proposed annotation system with a camera-only (CO) object 3D localization method \cite{wang2019monocular} and CRF algorithm \cite{wang2021rodnet}. Comparing with CO, the MAE achieved by our proposed annotator can be decreased by as much as $40\%$ after taking radar into consideration. 
% Differ from the radar, the faraway objects in the camera will not be suppressed by the reflections from the nearby objects, so that the detection recall of the camera-based method \cite{wang2019monocular} can be reached to more than $96\%$.
% Because our system can provide an accurate ground plane for each frame, those missing objects can also be projected from camera to radar. Therefore, our performance of the detection recall is also very good, achieving about $95\%$.
Besides, our system can achieve over $90\%$ for both the precision and recall. Overall, the DQF1 score of our system outperforms CO \cite{wang2019monocular} by about $15\%$. Comparing with CRF \cite{wang2021rodnet}, the MAE and precision are similar but the recall of CRF is much worse, which means that CRF has a number of false negatives. These false negatives may be due to the large errors from the results of the CO method. Overall, our annotator improves the DQF1 performance by around $6\%$, compared with CRF. 
The qualitative results are shown in Fig.~\ref{fig:results}, where the objects are accurately annotated even if they are not detected by the radar.

\begin{table}[t]
\begin{center}
\footnotesize
\begin{tabular}{L{1.3cm}|C{1.55cm}C{1cm}C{1cm}|C{1.1cm}}
\hline
Methods & MAE & Precision & Recall & \textbf{DQF1}\\
\hline
CO \cite{wang2019monocular} & $1.21$ ($\pm 1.05$) & $81.10\%$ & $96.11\%$ & $55.16\%$\\
CRF \cite{wang2021rodnet} & $0.68$ ($\pm 0.72$) & $93.02\%$ & $70.11\%$ & $64.14\%$ \\
\hline
\textbf{Ours} & $0.72$ ($\pm 0.78$) & $90.57\%$ & $95.35\%$ & $\mathbf{70.36\%}$\\
\hline
\end{tabular}
\end{center}
\caption{Performance evaluation of different annotation methods on the \textit{\textbf{selected training}} set (campus road and city street).}
\label{tab:results}
\end{table}

% \subsection{Ablation Studies}

% We analyze the performance of our system after each stage, i.e., detection alignment (DA), and supplementary projection (SP). The results are shown in Table~\ref{tab:ablation_stages}. The MAE is very low at the CA and FA stage because it just includes the errors of the aligned objects. While the recall is relatively poor since the unaligned objects are not annotated yet. After most objects detected by camera are projected radar in the SP stage, the recall is improved significantly. However, we also notice that the MAE increases and the precision drops a little bit in the SP stage. That is because the locations of projected objects are not as accurate as the aligned objects. It also means that there is a trade-off between MAE and recall. 

% \begin{table}[t]
% \begin{center}
% \footnotesize
% \begin{tabular}{C{1.2cm}|C{1.6cm}C{1cm}C{1cm}|C{1cm}}
% \hline
% Stages & MAE & Precision & Recall & DQF1\\
% \hline
% CA & $0.58$ ($\pm 0.63$) & $97.20\%$ & $50.74\%$ & $58.67\%$\\
% FA & $0.55$ ($\pm 0.56$) & $98.80\%$ & $60.28\%$ & $66.28\%$\\
% SP & $0.72$ ($\pm 0.78$) & $90.57\%$ & $95.35\%$ & $\mathbf{70.36\%}$\\
% \hline
% \end{tabular}
% \end{center}
% \caption{The performance of different stages in our system.}
% \label{tab:ablation_stages}
% \end{table}

We also use different time window sizes $t$ in the ground plane optimization (Eq.~\ref{eq:opt}) and evaluate the performance to choose a good time window $t$. The results are shown in Table~\ref{tab:ablation_win}. Besides, with larger window size $t$, the recall increases and the DQF1 score gradually converges. Overall, the best DQF1 score is achieved at $t=50$. 

\begin{table}[t]
\begin{center}
\footnotesize
\begin{tabular}{C{1.4cm}|C{1.6cm}C{1cm}C{1cm}|C{1.1cm}}
\hline
Window $t$ & MAE & Precision & Recall & \textbf{DQF1}\\
\hline
1 & $0.69$ ($\pm 0.77$) & $90.78\%$ & $91.03\%$ & $65.39\%$\\
5 & $0.71$ ($\pm 0.79$) & $90.70\%$ & $92.88\%$ & $66.22\%$\\
10 & $0.72$ ($\pm 0.80$) & $91.06\%$ & $93.47\%$ & $67.33\%$\\
50 & $0.72$ ($\pm 0.78$) & $90.57\%$ & $95.35\%$ & $\mathbf{70.36\%}$\\
100 & $0.73$ ($\pm 0.79$) & $90.34\%$ & $95.87\%$ & $70.35\%$\\
\hline
\end{tabular}
\end{center}
\caption{The performance using different time window sizes evaluated on the \textit{\textbf{selected training}} set (campus road and city street).}
\label{tab:ablation_win}
\end{table}

\section{Conclusion}
\label{sec:conclusion}

In this paper, we proposed a novel radar object detection platform for adverse driving scenarios, including a large-scale dataset, annotation and evaluation system. 
This platform is potentially valuable to the autonomous driving community for the deep learning based radar semantic understanding tasks, e.g., detection, segmentation, tracking, etc. It is also an inspiration for a new autonomous vehicle solution using a camera-radar sensor system for all-weather conditions. 

\section*{Acknowledgement}
This research work was partially supported by CMMB Vision -- UWECE Center on Satellite Multimedia and Connected Vehicles. The authors would also like to thank the colleagues and students in Information Processing Lab (IPL) at the University of Washington for their help and assistance on the dataset collection, processing, and annotation works.

{\balance \small
\bibliographystyle{ieee_fullname}
\bibliography{egbib}

\begin{thebibliography}{10}\itemsep=-1pt

\bibitem{flir}
Flir systems.
\newblock \url{https://www.flir.com/}.

\bibitem{ti}
Texas instruments.
\newblock \url{http://www.ti.com/}.

\bibitem{apollo_scape_dataset}
Apollo scape dataset.
\newblock \url{http://apolloscape.auto/}, 2018.

\bibitem{waymo_open_dataset}
Waymo open dataset: An autonomous driving dataset.
\newblock \url{https://www.waymo.com/open}, 2019.

\bibitem{8468324}
A. {Angelov}, A. {Robertson}, R. {Murray-Smith}, and F. {Fioranelli}.
\newblock Practical classification of different moving targets using automotive
  radar and deep neural networks.
\newblock {\em IET Radar, Sonar Navigation}, 12(10):1082--1089, 2018.

\bibitem{ansari2018earth}
Junaid~Ahmed Ansari, Sarthak Sharma, Anshuman Majumdar, J~Krishna Murthy, and
  K~Madhava Krishna.
\newblock The earth ain't flat: Monocular reconstruction of vehicles on steep
  and graded roads from a moving camera.
\newblock In {\em 2018 IEEE/RSJ International Conference on Intelligent Robots
  and Systems (IROS)}, pages 8404--8410. IEEE, 2018.

\bibitem{RadarRobotCarDatasetICRA2020}
Dan Barnes, Matthew Gadd, Paul Murcutt, Paul Newman, and Ingmar Posner.
\newblock The oxford radar robotcar dataset: A radar extension to the oxford
  robotcar dataset.
\newblock In {\em Proceedings of the IEEE International Conference on Robotics
  and Automation (ICRA)}, Paris, 2020.

\bibitem{nuscenes2019}
Holger Caesar, Varun Bankiti, Alex~H. Lang, Sourabh Vora, Venice~Erin Liong,
  Qiang Xu, Anush Krishnan, Yu Pan, Giancarlo Baldan, and Oscar Beijbom.
\newblock nuscenes: A multimodal dataset for autonomous driving.
\newblock {\em arXiv preprint arXiv:1903.11027}, 2019.

\bibitem{cai2018cascade}
Zhaowei Cai and Nuno Vasconcelos.
\newblock Cascade r-cnn: Delving into high quality object detection.
\newblock In {\em Proceedings of the IEEE conference on computer vision and
  pattern recognition}, pages 6154--6162, 2018.

\bibitem{dong2020probabilistic}
Xu Dong, Pengluo Wang, Pengyue Zhang, and Langechuan Liu.
\newblock Probabilistic oriented object detection in automotive radar.
\newblock In {\em Proceedings of the IEEE/CVF Conference on Computer Vision and
  Pattern Recognition Workshops}, pages 102--103, 2020.

\bibitem{duan2019centernet}
Kaiwen Duan, Song Bai, Lingxi Xie, Honggang Qi, Qingming Huang, and Qi Tian.
\newblock Centernet: Keypoint triplets for object detection.
\newblock In {\em Proceedings of the IEEE International Conference on Computer
  Vision}, pages 6569--6578, 2019.

\bibitem{feng2020deep}
Di Feng, Christian Haase-Sch{\"u}tz, Lars Rosenbaum, Heinz Hertlein, Claudius
  Glaeser, Fabian Timm, Werner Wiesbeck, and Klaus Dietmayer.
\newblock Deep multi-modal object detection and semantic segmentation for
  autonomous driving: Datasets, methods, and challenges.
\newblock {\em IEEE Transactions on Intelligent Transportation Systems}, 2020.

\bibitem{geiger2013vision}
Andreas Geiger, Philip Lenz, Christoph Stiller, and Raquel Urtasun.
\newblock Vision meets robotics: The kitti dataset.
\newblock {\em The International Journal of Robotics Research},
  32(11):1231--1237, 2013.

\bibitem{he2017mask}
Kaiming He, Georgia Gkioxari, Piotr Doll{\'a}r, and Ross Girshick.
\newblock Mask r-cnn.
\newblock In {\em Proceedings of the IEEE international conference on computer
  vision}, pages 2961--2969, 2017.

\bibitem{6042174}
S. {Heuel} and H. {Rohling}.
\newblock Two-stage pedestrian classification in automotive radar systems.
\newblock In {\em 2011 12th International Radar Symposium (IRS)}, pages
  477--484, Sep. 2011.

\bibitem{hsu2020traffic}
Hung-Min Hsu, Yizhou Wang, and Jenq-Neng Hwang.
\newblock Traffic-aware multi-camera tracking of vehicles based on reid and
  camera link model.
\newblock In {\em Proceedings of the 28th ACM International Conference on
  Multimedia}, pages 964--972, 2020.

\bibitem{levinson2011towards}
Jesse Levinson, Jake Askeland, Jan Becker, Jennifer Dolson, David Held, Soeren
  Kammel, J~Zico Kolter, Dirk Langer, Oliver Pink, Vaughan Pratt, et~al.
\newblock Towards fully autonomous driving: Systems and algorithms.
\newblock In {\em 2011 IEEE Intelligent Vehicles Symposium (IV)}, pages
  163--168. IEEE, 2011.

\bibitem{lin2017focal}
Tsung-Yi Lin, Priya Goyal, Ross Girshick, Kaiming He, and Piotr Doll{\'a}r.
\newblock Focal loss for dense object detection.
\newblock In {\em Proceedings of the IEEE international conference on computer
  vision}, pages 2980--2988, 2017.

\bibitem{major2019vehicle}
Bence Major, Daniel Fontijne, Amin Ansari, Ravi Teja~Sukhavasi, Radhika
  Gowaikar, Michael Hamilton, Sean Lee, Slawomir Grzechnik, and Sundar
  Subramanian.
\newblock Vehicle detection with automotive radar using deep learning on
  range-azimuth-doppler tensors.
\newblock In {\em Proceedings of the IEEE International Conference on Computer
  Vision Workshops}, 2019.

\bibitem{meyer2019automotive}
Michael Meyer and Georg Kuschk.
\newblock Automotive radar dataset for deep learning based 3d object detection.
\newblock In {\em 2019 16th European Radar Conference (EuRAD)}, pages 129--132.
  IEEE, 2019.

\bibitem{mousavian20173d}
Arsalan Mousavian, Dragomir Anguelov, John Flynn, and Jana Kosecka.
\newblock 3d bounding box estimation using deep learning and geometry.
\newblock In {\em Proceedings of the IEEE Conference on Computer Vision and
  Pattern Recognition}, pages 7074--7082, 2017.

\bibitem{murthy2017reconstructing}
J~Krishna Murthy, GV~Sai Krishna, Falak Chhaya, and K~Madhava Krishna.
\newblock Reconstructing vehicles from a single image: Shape priors for road
  scene understanding.
\newblock In {\em 2017 IEEE International Conference on Robotics and Automation
  (ICRA)}, pages 724--731. IEEE, 2017.

\bibitem{ouaknine2020carrada}
Arthur Ouaknine, Alasdair Newson, Julien Rebut, Florence Tupin, and Patrick
  P{\'e}rez.
\newblock Carrada dataset: Camera and automotive radar with range-angle-doppler
  annotations.
\newblock {\em arXiv preprint arXiv:2005.01456}, 2020.

\bibitem{palffy2020cnn}
Andras Palffy, Jiaao Dong, Julian~FP Kooij, and Dariu~M Gavrila.
\newblock Cnn based road user detection using the 3d radar cube.
\newblock {\em IEEE Robotics and Automation Letters}, 5(2):1263--1270, 2020.

\bibitem{redmon2018yolov3}
Joseph Redmon and Ali Farhadi.
\newblock Yolov3: An incremental improvement.
\newblock {\em arXiv preprint arXiv:1804.02767}, 2018.

\bibitem{ren2015faster}
Shaoqing Ren, Kaiming He, Ross Girshick, and Jian Sun.
\newblock Faster r-cnn: Towards real-time object detection with region proposal
  networks.
\newblock In {\em Advances in neural information processing systems}, pages
  91--99, 2015.

\bibitem{richards2005fundamentals}
Mark~A Richards.
\newblock {\em Fundamentals of radar signal processing}.
\newblock Tata McGraw-Hill Education, 2005.

\bibitem{schubert2017dbscan}
Erich Schubert, J{\"o}rg Sander, Martin Ester, Hans~Peter Kriegel, and Xiaowei
  Xu.
\newblock Dbscan revisited, revisited: why and how you should (still) use
  dbscan.
\newblock {\em ACM Transactions on Database Systems (TODS)}, 42(3):1--21, 2017.

\bibitem{sheeny2020radiate}
Marcel Sheeny, Emanuele De~Pellegrin, Saptarshi Mukherjee, Alireza Ahrabian,
  Sen Wang, and Andrew Wallace.
\newblock Radiate: A radar dataset for automotive perception.
\newblock {\em arXiv preprint arXiv:2010.09076}, 2020.

\bibitem{song2015joint}
Shiyu Song and Manmohan Chandraker.
\newblock Joint sfm and detection cues for monocular 3d localization in road
  scenes.
\newblock In {\em Proceedings of the IEEE Conference on Computer Vision and
  Pattern Recognition}, pages 3734--3742, 2015.

\bibitem{wang2019exploit}
Gaoang Wang, Yizhou Wang, Haotian Zhang, Renshu Gu, and Jenq-Neng Hwang.
\newblock Exploit the connectivity: Multi-object tracking with trackletnet.
\newblock In {\em Proceedings of the 27th ACM International Conference on
  Multimedia}, pages 482--490, 2019.

\bibitem{wang2019monocular}
Yizhou Wang, Yen-Ting Huang, and Jenq-Neng Hwang.
\newblock Monocular visual object 3d localization in road scenes.
\newblock In {\em Proceedings of the 27th ACM International Conference on
  Multimedia}, pages 917--925. ACM, 2019.

\bibitem{wang2021rodnet}
Yizhou Wang, Zhongyu Jiang, Xiangyu Gao, Jenq-Neng Hwang, Guanbin Xing, and Hui
  Liu.
\newblock Rodnet: Radar object detection using cross-modal supervision.
\newblock In {\em Proceedings of the IEEE/CVF Winter Conference on Applications
  of Computer Vision}, pages 504--513, 2021.

\bibitem{wang2021rodnet2}
Yizhou Wang, Zhongyu Jiang, Yudong Li, Jenq-Neng Hwang, Guanbin Xing, and Hui
  Liu.
\newblock Rodnet: A real-time radar object detection network cross-supervised
  by camera-radar fused object 3d localization.
\newblock {\em IEEE Journal of Selected Topics in Signal Processing}, 2021.

\bibitem{yang2021fast}
Hao Yang, Chenxi Liu, Meixin Zhu, Xuegang Ban, and Yinhai Wang.
\newblock How fast you will drive? predicting speed of customized paths by deep
  neural network.
\newblock {\em IEEE Transactions on Intelligent Transportation Systems}, 2021.

\bibitem{yang2020hybrid}
Hao Yang, Chenxi Liu, Meixin Zhu, Wei Sun, and Yinhai Wang.
\newblock Hybrid data-fusion model for short-term road hazardous segments
  identification based on the acceleration and deceleration information.
\newblock In {\em International Conference on Transportation and Development
  2020}, pages 313--326. American Society of Civil Engineers Reston, VA, 2020.

\bibitem{yang2020novel}
Hao~Frank Yang.
\newblock {\em Novel Traffic Sensing Using Multi-Camera Car Tracking and
  Re-Identification (MCCTRI)}.
\newblock PhD thesis, 2020.

\end{thebibliography}
}

\end{document}